\documentclass[a4paper]{jpconf} 
\usepackage{graphicx}
\usepackage{cleveref}
\usepackage{lipsum}

\renewcommand{\thefootnote}{}
\begin{document}
\title{Precise Apple Detection and Localization in Orchards using YOLOv5 for Robotic Harvesting Systems}

\author{Ziyue Jiang$^{1}$, 
Bo Yin$^{1,*}$ and 
Boyun Lu$^{1}$}


\address{\textbf{These authors contributed equally to this work}}
\address{School of Future Technology, South China University of Technology, Guangzhou, China}

\ead{202130200293@mail.scut.edu.cn \\
\hspace{3.5em}$^{*}$202130300771@mail.scut.edu.cn \\
\hspace{3.5em}202130191508@mail.scut.edu.cn}
\begin{abstract}
The advancement of agricultural robotics holds immense promise for transforming fruit harvesting practices, particularly within the apple industry.   The accurate detection and localization of fruits are pivotal for the successful implementation of robotic harvesting systems.   In this paper, we propose a novel approach to apple detection and position estimation utilizing an object detection model, YOLOv5.   Our primary objective is to develop a robust system capable of identifying apples in complex orchard environments and providing precise location information.   To achieve this, we curated an autonomously labeled dataset comprising diverse apple tree images, which was utilized for both training and evaluation purposes. Through rigorous experimentation, we compared the performance of our YOLOv5-based system with other popular object detection models, including SSD.   Our results demonstrate that the YOLOv5 model outperforms its counterparts, achieving an impressive apple detection accuracy of approximately 85\%.  We believe that our proposed system's accurate apple detection and position estimation capabilities represent a significant advancement in agricultural robotics, laying the groundwork for more efficient and sustainable fruit harvesting practices.
\end{abstract}

\section{Introduction}
The rapid advancement of Artificial Intelligence (AI) has spurred the development of cutting-edge technologies in the field of agriculture, such as precision and smart farming. Techniques of agricultural computer vision are revolutionizing modern agriculture by enabling farmers to effectively monitor crops and enhance yields. Nevertheless, there remain substantial challenges in creating accurate and efficient fruit inspection systems for harvesting robots, including issues with occlusions, varying lighting conditions, and the necessity for real-time performance capabilities. Previous studies have investigated traditional image processing and deep learning approaches; However, these methods still cannot fully meet the needs of picking robots in actual working environments. Moreover, agricultural harvesting entails not only the identification of crops but also the ascertainment of crop distribution and center coordinates. To augment the efficacy of harvesting robots, an appropriate model and mathematical methods are required to calculate the distribution of crops and their central coordinates. YOLO v5, an advanced object detection algorithm\cite{ge2021yolox}, has demonstrated exceptional performance in image recognition and is capable of precisely localizing objects with bounding boxes, presenting itself as a promising solution for these challenges.

\section{Exploring the Inner Workings of YOLOv5}
\subsection{Model Overview}
YOLOv5 is an object detection model and the latest iteration in the YOLO\cite{redmon2016you} (You Only Look Once) series. Developed by the Ultralytics team, it builds upon the YOLOv4 architecture with improvements and optimizations, offering higher detection accuracy and faster inference speed.

The following are several notable characteristics and improvements incorporated into the YOLOv5 object detection algorithm:
\begin{enumerate}
    \item \textbf{Lightweight model:} YOLOv5 achieves high accuracy while being lightweight, enabling real-time detection on embedded devices and mobile platforms through model optimization and pruning.
    
    \item \textbf{Improved backbone network:} YOLOv5 utilizes a new backbone network\cite{li2018detnet} called CSPDarknet53, which outperforms Darknet53 used in YOLOv4 in terms of performance and speed.
    
    \item \textbf{Data augmentation and training techniques:} YOLOv5 employs efficient data augmentation techniques such as mosaic data augmentation\cite{dadboud2021single} and multi-scale training\cite{singh2018sniper} to enhance model generalization and robustness.
    
    \item \textbf{Automatic model optimization:} YOLOv5 introduces automatic model optimization techniques that adjust model architecture and hyperparameters automatically to further enhance model performance.
\end{enumerate}

\subsection{Model Structure}
\begin{figure}[htbp]
    \centering
    \includegraphics[width=\textwidth]{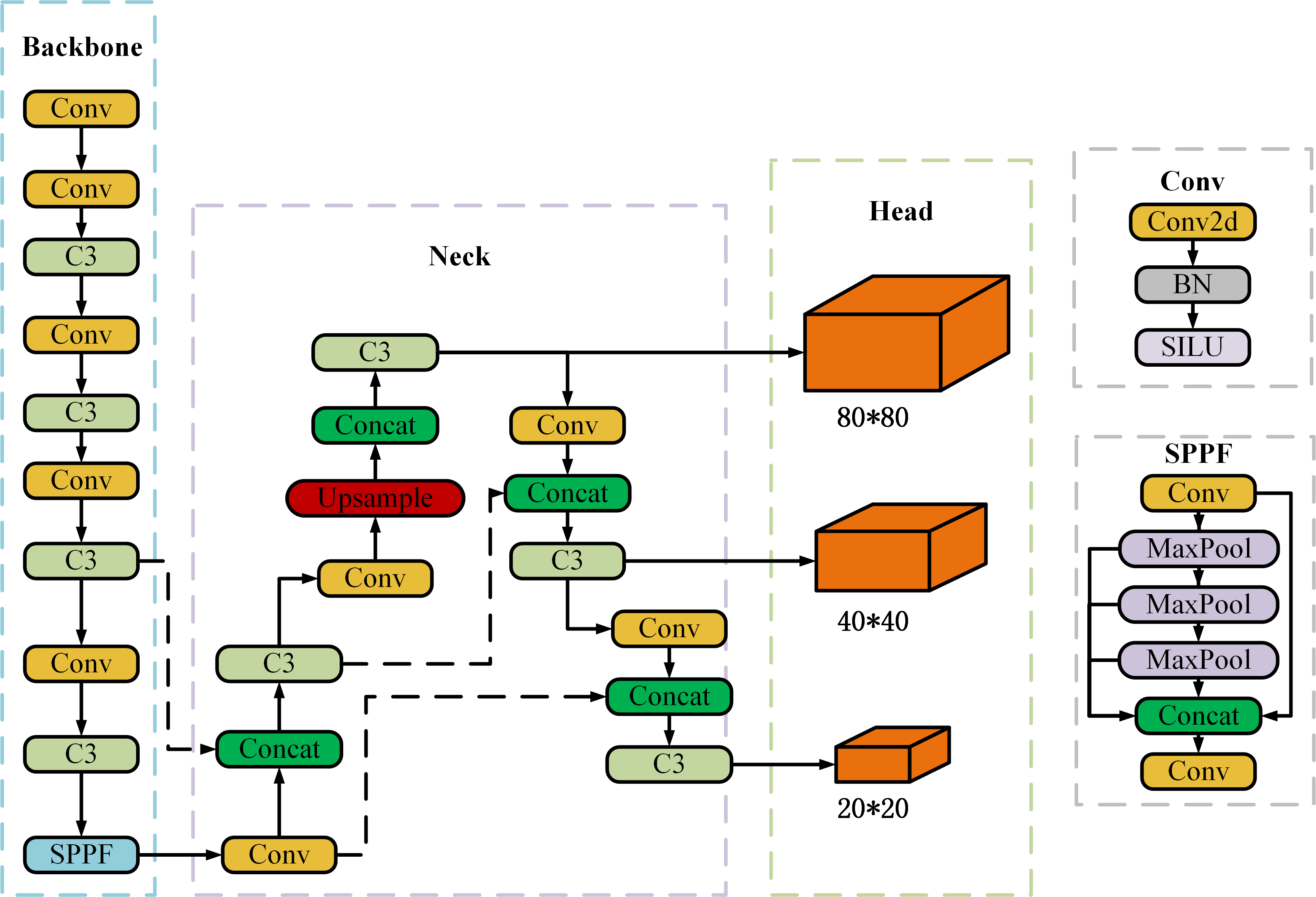}
    \caption{YOLOv5 Structure}
    \label{yolov5s}
\end{figure}

Figure \ref{yolov5s} illustrates the basic architecture of the YOLOv5 object detection model. The entire model can be divided into three main parts: the backbone network, the neck network, and the detection head.

The backbone network typically consists of convolutional layers (Conv) and residual blocks (C3)\cite{he2016deep}, used for extracting image features. Features are extracted and fused layer by layer in the backbone network.

The neck part utilizes the feature pyramid network (FPN) structure, combining features from different levels through upsampling and concatenate operations, forming feature maps with rich semantic information.

The detection head part includes a series of convolutional layers and feature fusion operations. It ultimately outputs three feature maps at different scales (80x80, 40x40, 20x20), used for predicting bounding boxes and class information for objects of different sizes. The SPFF module is introduced during the detection process for spatial pyramid pooling, assisting in detecting multi-scale objects.

Overall, the YOLOv5 model adopts a design with a backbone network for feature extraction, a neck network for feature fusion, and a detection head for object detection. Through effective feature fusion and pyramid structures, it achieves good detection accuracy and real-time performance in single-stage detection tasks.

\section{Experiment}

\subsection{Data Collection}
To facilitate the development and assessment of the YOLOv5-based system for apple detection, a comprehensive dataset of apple images has been compiled. This dataset was curated through a combination of web scraping methodologies and utilization of an in-house image repository. Emphasis was placed on acquiring images that depict a diverse array of apple varieties, orchard environments, lighting conditions, and degrees of occlusion, thereby augmenting the model's capacity for generalization. Subsequent to the acquisition of raw image data, a meticulous manual annotation procedure was conducted to precisely delineate bounding boxes around individual apples within each image. These annotations involved the assignment of appropriate class labels to facilitate subsequent model training and evaluation.

\subsection{Experiment Setting}
The experimentation utilized a dataset comprising orchard fruit trees, meticulously annotated with the locations of apples. This dataset was partitioned into distinct training, validation, and test sets, adhering to a proportional ratio of 3:1:1.

In the pursuit of optimizing the YOLOv5m architecture for the specific task of apple recognition, careful consideration was given to its inherent balance between computational efficiency and performance. A tailored approach involving model fine-tuning was employed, leveraging a pre-training strategy on our comparatively constrained dataset. Augmenting the dataset with diverse transformations, such as random rotation, cropping, and zooming, was instrumental in augmenting the model's capacity for generalization.

Moreover, refinement of the YOLOv5's anchor boxes was undertaken to better align with the size and morphology of the dataset under scrutiny. Implementation of a fine-tuned learning rate regimen, complemented by a decay strategy, aimed to foster training stability and convergence. Notably, meticulous adjustments to the weighting of various loss components within the loss function were executed to ensure a harmonized influence throughout the training trajectory.
\subsection{Experiment Result}
\begin{figure}[htbp]
    \centering
    \includegraphics[width=\textwidth]{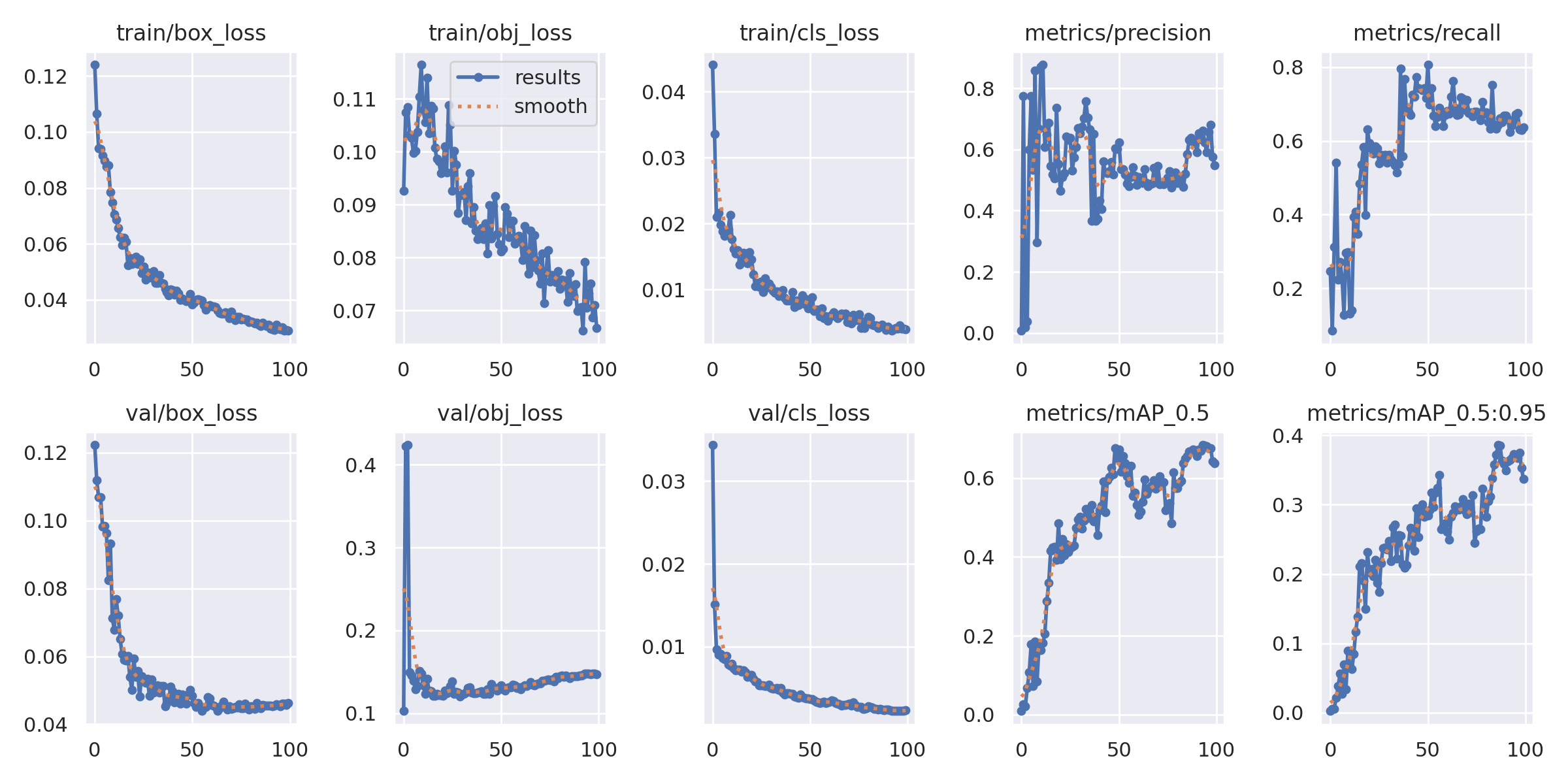}
    \caption{Various Indicators During the Training of the YOLOv5 Model}
    \label{yolov5}
\end{figure}

Figure \ref{yolov5} depicts the progression of diverse indicators throughout the training of the YOLOv5 model. These curves delineate various types of loss and performance metrics crucial for assessing and monitoring the model's efficacy during training.

As the training progresses, both the training and validation losses exhibit a diminishing trend, indicating an improvement in the model's ability to minimize errors on both datasets. Concurrently, the performance metrics exhibit an overall ascending trajectory with fluctuations, suggesting the model's gradual enhancement in object detection proficiency throughout training.

Figures 3 and 4 present the apple recognition outcomes across various scenarios. Notably, on the test set, our model demonstrates an accuracy of 85\%, underscoring its effectiveness in accurately identifying apples across diverse conditions.
\begin{figure}[htbp]
    \centering
    \begin{minipage}{.5\textwidth}
      \centering
      \includegraphics[width=0.9\linewidth]{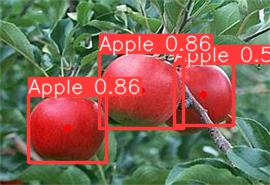}
      \label{sub1}
      \caption{Extract big apples}
    \end{minipage}%
    \begin{minipage}{.5\textwidth}
      \centering
      \includegraphics[width=0.9\linewidth]{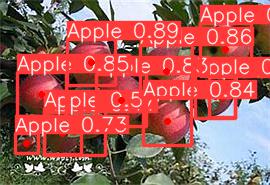}
      \label{sub2}
      \caption{Extract small apples}
    \end{minipage}
\end{figure}

\subsection{Algorithm Comparison}
Furthermore, an extensive comparison was conducted between YOLOv5 and SSD\cite{liu2016ssd} to delineate their suitability for the specific task of apple picking. As delineated in Table \ref{comp}, it is evident that YOLOv5 outperforms SSD across multiple metrics. Notably, YOLOv5 demonstrates superior speed and model size efficiency compared to SSD, thus rendering it more adept for real-time applications. Moreover, in terms of accuracy, while YOLOv5 marginally surpasses SSD, further affirming its efficacy in precise object detection tasks such as apple picking.

In conclusion, YOLOv5's advanced technology and exceptional performance in real-time object detection, particularly its promising applications in agricultural image processing, position it as an ideal solution for the task of apple detection and localization.  Its ability to swiftly and accurately identify objects aligns well with the complexities inherent in agricultural environments, rendering it a valuable tool for enhancing productivity and efficiency in apple-related processes.
\begin{center}
\begin{table}[!t]
\caption{YOLOv5 vs SSD}
\centering
\begin{tabular}{@{}l*{7}{l}}
\br
Algorithm&Speed(on Tesla V100)&mAP(on COCO)&Size\\
\mr
SSD&\hspace{3em}59FPS &\hspace{3em}55.1&34MB\\
YOLOv5 &\hspace{3em}140FPS&\hspace{3em}57.4&24MB\\
\br
\end{tabular}
\label{comp}
\end{table}
\end{center}

\section{Challenge and Prospect}
Over time, the YOLO algorithm has undergone successive iterations, yielding diverse versions optimized for distinct tasks and data attributes. In agricultural settings, where precise, swift, real-time, and resilient object detection is imperative, YOLO emerges as a fitting solution. It excels in striking a delicate balance between computational efficiency and detection accuracy. Noteworthy is YOLO's capability to swiftly and resource-efficiently discern objects in real-time scenarios, while still maintaining commendable accuracy.  For example, Using YOLO+MRM\cite{ye2020experimental},  detected over 180,000 broilers per hour in an automatic broiler-slaughter line.
Nasirahmadi demonstrated that YOLOv4 achieved a frame rate of 29 fps while maintaining accuracy above 90\% for mechanical sugar beet damage detection during harvesting.

The performance of object detection in agriculture is significantly influenced by the variability of agricultural objects and environmental complexity. Therefore, researchers frequently underscore the importance of assembling extensive and varied datasets for model training, encompassing diverse growth stages, lighting conditions, and multiple field locations or scenarios, to develop models that are both accurate and robust. It is imperative to collect a wide range of training images to account for environmental variability, including variations in lighting conditions, seasonal changes, and weather conditions such as rain, clouds, and brightness.

\section{Conclusion}
This paper delves into the identification and detection of apples within orchard scenes utilizing YOLOv5. Achieving an 85\% accuracy rate on specifically curated datasets, it showcases the efficacy of the approach.  Furthermore, the paper elucidates the suitability of YOLOv5 for orchard environments, drawing comparisons with the SSD algorithm to underscore its advantages.  In conclusion, the paper provides a summary of the findings and anticipates the future prospects and challenges inherent in natural scene analysis.  Additionally, it offers insights into potential directions for further improvement in this field.



\newpage
\section*{References}
\bibliographystyle{unsrt}
\bibliography{ref}
\end{document}